\pdfoutput=1

\documentclass[11pt]{article}

\usepackage[final]{acl}

\usepackage{times}
\usepackage{latexsym}

\usepackage[T1]{fontenc}
\usepackage{booktabs}

\usepackage[utf8]{inputenc}

\usepackage{microtype}

\usepackage{inconsolata}

\usepackage{graphicx}

\usepackage{amsmath}
\usepackage{amssymb}
\usepackage{mathtools}
\usepackage{amsthm}
\usepackage{algorithm}
\usepackage[most]{tcolorbox}
\usepackage{colortbl}
\usepackage{soul}     
\usepackage{xcolor}   
\usepackage{multirow}
\usepackage{multicol}
\definecolor{grey}{gray}{0.85}
\usepackage{color}
\usepackage{enumitem}
\usepackage{bbm}
\definecolor{lightred}{rgb}{1, 0.8, 0.8}
\definecolor{lightgreen}{rgb}{0.8, 1, 0.8}
\definecolor{lightblue}{rgb}{0.8, 0.9, 1}
\newcommand*\colourcheck[1]{%
  \expandafter\newcommand\csname #1check\endcsname{\textcolor{#1}{\ding{52}}}%
}
\definecolor{rowgray}{gray}{0.95}
\definecolor{headergray}{gray}{0.85}
\usepackage{algpseudocode}
\colourcheck{blue}
\colourcheck{green}
\colourcheck{red}
\usepackage{pifont}
\newcommand{\xmark}{\ding{55}}%

\renewcommand{\arraystretch}{1.3}

\DeclareMathOperator*{\argmax}{arg\,max}

\usepackage[capitalize,noabbrev]{cleveref}
\theoremstyle{plain}

\theoremstyle{definition}

\theoremstyle{remark}

\usepackage{booktabs}
\usepackage{multirow}
\usepackage{colortbl}
\usepackage{caption}
\usepackage{subcaption}
\usepackage{geometry}
\usepackage{bm}
\definecolor{rowgray}{gray}{0.95}
\definecolor{headergray}{gray}{0.85}

\usepackage{makecell}
\usepackage{titlesec}

\title{COCO-Tree: \underline{CO}mpositional Hierarchical \underline{CO}ncept \underline{Tree}s for Enhanced Reasoning in Vision Language Models}

\author{
  Sanchit Sinha \and Guangzhi Xiong \and Aidong Zhang \\
  University of Virginia \\
  Charlottesville, VA, USA \\
  \texttt{\{sanchit, hhu4zu, aidong\}@virginia.edu}
}

\begin{document}
\maketitle

\begin{abstract}
Compositional reasoning remains a persistent weakness of modern vision language models (VLMs): they often falter when a task hinges on understanding how multiple objects, attributes, and relations interact within an image. Multiple research works have attempted to improve compositionality performance by creative tricks such as improving prompt structure, chain of thought reasoning, etc. A more recent line of work attempts to impart additional reasoning in VLMs using well-trained Large Language Models (LLMs), which are far superior in linguistic understanding than VLMs to compensate for the limited linguistic prowess of VLMs. However, these approaches are either resource-intensive or do not provide an interpretable reasoning process. In this paper, we present ``COCO-Tree'' - a novel approach that augments VLM outputs with carefully designed neurosymbolic concept trees learned from LLMs to improve VLM's linguistic reasoning. COCO-Tree's beam search-inspired reasoning process boosts compositionality performance and provides a rationale behind VLM predictions. Empirical results on four compositionality benchmarks, Winoground, EqBench, ColorSwap, and SugarCrepe, in seven different open-source VLMs with varying sizes, demonstrate that COCO-Tree significantly improves compositional generalization by 5-10\% over baselines.
The code is available at: \url{https://github.com/sanchit97/compositionality-low-res-vlm}
\end{abstract}
\section{Introduction}
\label{sec:intro}
Vision Language Models (VLMs) \cite{radford2021learning} have achieved state-of-the-art performance in complex tasks such as video understanding \cite{tang2024video} and scene captioning \cite{li2024wolf}. Significant research improves upon their performance by proposing newer architectures \cite{maniparambil2023enhancing}, better pre-training paradigms \cite{castro2024clove}, etc. In parallel, many fundamental challenges have been identified in VLMs, such as a lack of numerical reasoning \cite{zhang2020vinvl}, limited spatial reasoning \cite{kamath2023whats}, hallucinated outputs \cite{ji2023survey}, etc. Some studies \cite{zeng2024investigating,hua2024mmcomposition,ma2022learning} show that many VLMs behave like a `bag of visual words' rather than true reasoners \cite{doveh2023teaching,dumpala2024seeing,herzig2023incorporating}. This gap limits deployment in safety-critical domains such as medical imaging and industrial safety. 

\begin{figure}[t]
    \centering
    \includegraphics[width=0.48\textwidth]{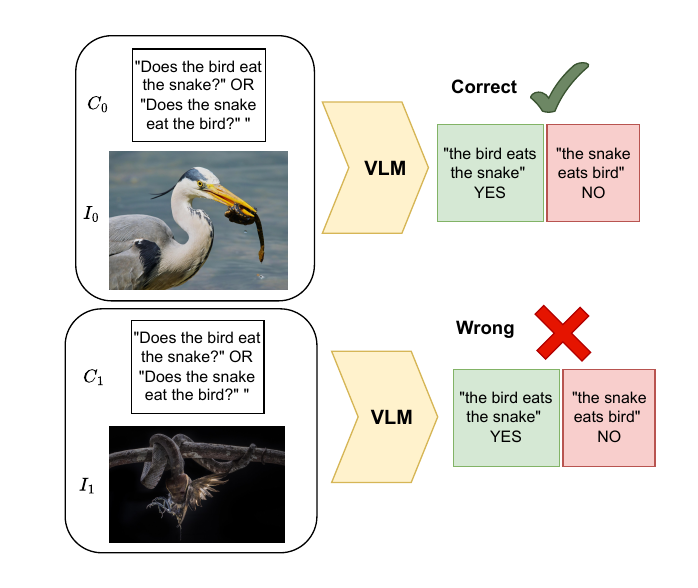}
    \caption{\small An example of the measure of compositionality problem from the Winoground dataset. The VLM is successful in identifying the presence of a bird and snake in the image but is unable to correctly \textit{understand} the relations between them.}
    \label{fig:motivation}
\end{figure}

One such critical problem is \textbf{compositionality} - wherein the model successfully identifies objects and attributes in an image but fails to accurately understand the relationships \underline{between} them. For example, in Figure~\ref{fig:motivation}, a VLM can correctly identify a bird and a snake in the image, but cannot differentiate between simple semantic questions: `Does the bird eat the snake?' and `Does the snake eat the bird?'. Compositionality has been a high research activity \cite{zeng2024investigating,hua2024mmcomposition,ma2022learning} and remains a challenging frontier.

A possible reason for the lackluster performance of compositionality is a lack of robust linguistic reasoning processes in VLMs. VLM pre-training is highly dependent on fine-tuning image-caption pairs, which causes catastrophic forgetting of linguistic reasoning and large distribution shifts \citep{zhai2023investigating}. This observation is validated by the fact that similar-sized LLMs, on which VLMs are built, often significantly outperform VLMs in language understanding \cite{wang2024picture}. Some recent approaches have used creative prompting techniques \cite{mitra2024compositional} to generate structured intermediate outputs (e.g., scene graphs, relationship ontology, etc.) that can guide VLMs towards better compositionality performance. Some approaches manually decompose captions into easily understandable entities that are easier for VLMs to understand \cite{cho2023davidsonian}. However, due to the limited linguistic expressibility of VLMs, standalone VLMs are unable to be truly effective at compositionality.

Using a strong LLM reasoner to augment the lack of linguistic reasoning process of VLMs can significantly improve compositionality. Some approaches preprocess the textual input \cite{cho2023davidsonian,hu2023tifa} while other approaches postprocess the VLM outputs to generate the desired output \cite{lin2023revisiting}. Most of these approaches focus on utilizing state-of-the-art LLMs to impart linguistic reasoning in VLMs, which require significantly higher resources as compared to the VLMs themselves. In addition, these approaches are often not directly interpretable or heuristic, hindering widespread adoption. We aim to design an approach that can effectively impart linguistic reasoning in VLMs using an external LLM of a similar scale and also provides an interpretable pathway through symbolic concepts. 

In this paper, we propose a novel approach ``COCO-Tree: \underline{CO}mpositional Hierarchical \underline{CO}ncept \underline{Tree}s'', which recursively decomposes textual inputs to VLMs into structurally similar but semantically different morphological entities, which are further used to learn associated neurosymbolic concept trees with an LLM reasoner. COCO-Tree further employs a novel beam-search-inspired path-finding strategy by exploring the learned concept trees. Finally, COCO-Tree augments the VLM outputs with the neurosymbolic learned path concepts, which not only improves compositionality performance but also provides a reasoning \textit{rationale} (concepts along the path) for VLMs. Our approach can be thought of as imparting System-2 \cite{nye2021improving} (slow, logical) reasoning through concept tree exploration into System-1 (fast, opaque) predictions. Specifically, our contributions are:
\begin{itemize}[leftmargin=*, parsep=0pt, itemsep=0pt, topsep=0pt]
    \item We propose COCO-Tree - a novel approach that creates hierarchical concept trees associated with textual inputs and subsequently finds reasoning pathways to augment VLM outputs.
    \item COCO-Tree is evaluated on four benchmark datasets, Winoground, EqBench, SugarCrepe, and ColorSwap, in seven open-source VLMs, resulting in a 5-10\% increase in compositionality performance compared to baselines.
    \item We conduct extensive ablation studies to validate the effect of each component of COCO-Tree and propose two novel path-finding strategies based on greedy and beam search.
    \item We demonstrate through a strong LLM reasoner that neurosymbolic reasoning pathways discovered with COCO-Tree improve interpretability. 
\end{itemize}



\section{Related Work}
\noindent \textbf{Compositionality Problem in VLMs.} Compositionality \cite{zeng2024investigating,hua2024mmcomposition,ma2022learning} remains a challenging problem for VLMs. Multiple works such as \citet{doveh2023teaching,herzig2023incorporating,dumpala2024seeing} show VLMs are barely better than object detectors. Collectively, these studies underscore the importance of developing and implementing strategies to enhance compositional reasoning in VLMs, ensuring a more accurate and nuanced understanding of complex visual and textual information.

\noindent\textbf{LLMs as Strong Reasoners to Augment VLMs.} Many recent research approaches have focused on the integration of LLMs to enhance the reasoning capabilities of VLMs due to their inherent lack of linguistic understanding. Some approaches like \citet{cho2023davidsonian,hu2023tifa} impart language context to VLM inference in the form of scene graphs or language priors. Some approaches such as \citet{zhou2023vicor} impart LLM outputs in visual understanding.

\noindent \textbf{Hierarchical Concept Learning.} More recently, research has focused on structuring concepts from most abstract to least abstract. This idea was first proposed in \citet{panousis2023hierarchical}, and subsequently discussed in \citet{pittino2023hierarchical, liu2023primenet, pham2024peeb, sun2024eliminating}.

\noindent\textbf{Neurosymbolic Reasoning.} A parallel line of research has attempted to explain Neural Networks using propositional logic. Early works utilized specialized architectures and regularization like \citet{riegel2020logical, dong2019neural}. More recently, with the taxonomical classification of neurosymbolic systems \cite{nye2021improving}, many works have attempted to integrate System-1 and System-2 reasoning together such as \citet{saha2024system, wu2024symbol}. Lastly, some approaches have attempted to combine concept-based explanations and neurosymbolic reasoning \cite{barbiero2023interpretable}.

\noindent\textbf{Comparison to Related Work.}
Recent efforts to address the compositionality gap in VLMs fall into two resource–motivated settings. (1) \emph{Single-model} approaches rely \emph{solely} on a frozen VLM and squeeze more signal out of its own token probabilities via clever prompting or discriminative scoring.
VQAScore~\cite{lin2025evaluating} re-ranks answer candidates by treating the VLM as a binary classifier, and CCoT~\cite{mitra2024compositional} first asks the same VLM to output a scene graph and then feeds that graph back into a second prompt.
While inexpensive, these methods are limited by the VLM's fixed linguistic reasoning capacity and yield only flat, non-hierarchical rationales.
(2) \emph{Multi-model} approaches enlist \emph{external} LLMs or auxiliary VLMs to perform symbolic reasoning on top of the base VLMs. DSG~\cite{cho2023davidsonian} feeds a full scene graph into a large LLM for compositional queries. The most direct comparison to our approach is CECE~\cite{cascante2024natural}, which pairs a large LLM to refine the caption and generate a set of positive (entailments) and negative (contradiction) candidate phrases to augment reasoning. Our method and CECE operate in a different setting. CECE assumes access to large resources during inference and thus utilizes a strong LLM reasoner and a strong VLM to score the candidates in addition to the base VLM. Our approach assumes a more resource-constrained setting where during inference, only a similarly sized LLM is available in addition to the base VLM. In Table~\ref{tab:comparision}, we contrast related approaches based on if they are interpretable (Column 2), utilize an external LLM (Column 3), `Setting' (Column 4), and additional resources needed at inference time to achieve benchmark performance (Column 5).
\begin{table}[ht]
\centering
\renewcommand{\arraystretch}{1.2} 
\setlength{\tabcolsep}{12pt} 
\resizebox{0.49\textwidth}{!}{
\begin{tabular}{ccccc}
\toprule
\textbf{Method} & \textbf{Int?} & \textbf{LLM?} & \textbf{Setting} & \textbf{Resources} \\
\midrule
VQAScore \cite{lin2025evaluating} & \xmark & \xmark & Single & None\\
CCoT \cite{lin2025evaluating} & \xmark & \xmark & Single & None \\
DSG \cite{cho2023davidsonian} & \xmark & \redcheck & Multiple & LLM \\
CECE \cite{cascante-bonilla2025natural} & \redcheck & \redcheck & Multiple & LLM, VLM\\
COCO-Tree (Ours) & \redcheck & \redcheck & Multiple & LLM\\
\bottomrule
\end{tabular}}
\caption{Comparison between related approaches based on interpretability (Int?), LLM usage, Setting, and Resources required during inference.}
\label{tab:comparision}
\end{table}


\section{Methodology}
\label{sec:method}
In this section, we introduce our proposed approach - COCO-Tree. Our approach augments the standard VLM inference (System-1) with a robust neurosymbolic reasoning system (System-2). The reasoning system first constructs extensive concept trees, rooted at the associated candidate captions. Subsequently, the System-2 prediction is calculated using a novel path-finding mechanism that searches for the ideal neurosymbolic reasoning pathway through the concept tree. The final System-2 prediction is fused adaptively with System-1 prediction. 

\begin{figure*}[t]
    \centering
    \includegraphics[width=\textwidth]{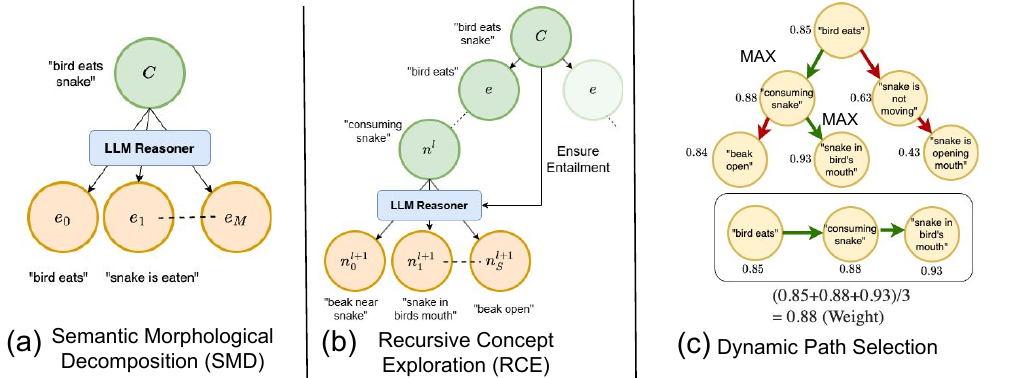}
    \caption{\small A schematic approach demonstrating the major components of our proposed approach. (a) Semantic Morphological Decomposition which decomposes a caption into morphological entities to disentangle structure and semantics. (b) Process of Recursive Concept Exploration, wherein new concepts are discovered. (c) Dynamic Path Selection and implied Neurosymbolic reasoning pathways. The numbers represent the composite scores and the green arrows represent the reasoning path selected.}
    \label{fig:schematic}
\end{figure*}

\subsection{Preliminaries and Notations}
\label{sec:problem-setting}
\noindent\textbf{Problem Setting.}
We first formally define the compositionality problem as represented in Figure~\ref{fig:motivation}. Let $\mathcal{I}$ denote the space of images and $\mathcal{C}$ the space of associated captions. We represent a VLM matching function as $f: \mathcal{I} \times \mathcal{C} \rightarrow \mathbb{R}$,
\noindent where $f(I, C)$ denotes an alignment score between image $I$ and caption $C$. Given a pair of images $\{I_0, I_1\}$ and a pair of associated captions $\{C_0, C_1\}$, the compositionality problem is characterized by three subtasks: Text, Image, and Group based on the binary indicator function ($\mathbbm{1}$) described below.

\noindent \textbf{(a) Text Task}. Given an image $I_t \in \{I_0, I_1\}$, Text score is given as $\mathbbm{1}[\hat{C}_t = C_t]$, where
\begin{equation}
    \hat{C}_t = \argmax_{C_i \in \{C_0, C_1\}} f(I_t, C_i)
\end{equation}
\noindent \textbf{(b) Image Task.}
Similarly, given a caption $C_t \in \{C_0, C_1\}$, Image score is given as $\mathbbm{1}[\hat{I}_t = I_t]$:
\begin{equation}
    \hat{I}_t = \argmax_{I_i \in \{I_0, I_1\}} f(I_i, C_t)
\end{equation}
\noindent \textbf{(c) Group Task.}
Based on Text and Image tasks, the group score is calculated on text/image scores:
\begin{equation}
 \mathbbm{1}[\hat{C}_t = C_t] \oplus \mathbbm{1}[\hat{I}_t = I_t]
\end{equation}
where $\oplus$ is the binary $AND$ operation.



\noindent\textbf{Hierarchical Concept Tree Notation.}
A conceptual structure encodes probabilistic semantics through node weights. Let a concept tree be defined as a rooted, directed, and acyclic graph:
\begin{equation}
    \mathcal{T} = (\mathcal{V}, \mathcal{E}, {C}_S), ~~{C_S}: \mathcal{V} \rightarrow [0, 1]
\end{equation}
where
$ \mathcal{V} = \{ v_1, v_2, \ldots, v_n \}$ is a finite set of nodes representing semantic concepts;
$\mathcal{E} \subseteq \mathcal{V} \times \mathcal{V}$, where each edge $(v_i, v_j) \in \mathcal{E}$ denotes a \textit{semantic entailment} (relation) from parent $v_i \in \mathcal{V}$ to child $v_j$ $ \in \mathcal{V}$; and $C_S$ is a node weighting function that assigns to each node $v \in \mathcal{V}$ a composite relevance score based on visual and linguistic grounding. Each concept tree associated with a caption $C$ is rooted at $C$ and is denoted as $\mathcal{T}_C$. 


\subsection{Concept Tree Construction}
Constructing the hierarchical concept tree for System-2 reasoning consists of three distinct steps. First, the initial caption is decomposed into disentangled morphological entities computed using a Semantic Morphological Decomposition function. Next, the tree is populated using a recursive concept exploration mechanism, which discovers representative visual concepts. Subsequently, each concept discovered is assigned a novel composite vision-language score based on its relevance to the image and the disentangled morphological entity. 

\noindent \textbf{Morphological Entities} are structurally discrete, non-overlapping phrases extracted from the caption that factor the sentence into multiple semantically different units - each representing a distinct part of the sentence. 

\noindent\textbf{Semantic Morphological Decomposition (SMD).}
A well-documented limitation of VLMs is their insufficient linguistic reasoning capabilities \cite{dumpala2024seeing}. In particular, conventional VLMs often struggle to disentangle linguistic \textit{structure} from \textit{semantics}. To address this, we introduce a morphological decomposition function $F_{SMD}$ characterized by a caption $C \in \mathcal{C}$, which partitions $C$ into discrete entities that are structurally distinct, but individually preserve the overall semantic meaning of the original caption. Mathematically, morphological entities $E \subset \mathcal{V}$ are calculated as:
\begin{equation} 
E  = F_{SMD}(C),~\text{where}~E = \{e_i\}_{i=1}^M 
\end{equation}
where $M$ is a tunable hyperparameter. Each $e_i \in E$ is a morphological entity, typically corresponding to a self-contained semantic concept, as shown in Figure~\ref{fig:schematic}(a). We use $e$ without index to denote each entity in the following paragraphs for brevity. Considering the example in Figure~\ref{fig:motivation}, the entities computed are ``bird eats" and ``snake gets eaten''. The set $E$ forms the initial nodes of the subsequent recursive concept exploration process.

\noindent\textbf{Recursive Concept Exploration (RCE).}
Starting with each initial morphological node, the linguistic reasoner recursively computes semantically similar visual concepts that are entailed by the root nodes. Intuitively, this function discovers a concept that is hierarchically related to its parent but is still entailed by the root caption. The recursive algorithm is akin to the Breadth-First Search (BFS) way of graph exploration, as shown in Figure~\ref{fig:schematic}(b). We denote the depth of the tree as $L$. For a given tree level $l<L$, the set of concept nodes at that level is signified as $N^l$. Subsequently, the set of concept nodes for the next level $l+1$, signified as $N^{l+1}$ can be mathematically computed as:
\begin{equation}
     N^{l+1} = \bigcup_{n^l_i \in N^l}  F_{RCE}(n^l_i, ~C, ~S)
\end{equation}
where the function $F_{RCE}$ is characterized by a concept node $n^l_i$, the caption $C$, and splitting factor $S$.  $S$ denotes the number of child nodes discovered at each level during recursive concept tree formation and controls the maximum number of concepts to discover per node.

\noindent\textbf{Composite Vision-Language Score.}
The constructed concept tree spans a multitude of diverse visual concepts organized in a hierarchy. Even though multiple concepts can be simultaneously relevant for a given image, the vast majority of discovered concepts are irrelevant to understanding compositionality. Hence, we utilize a scoring metric to assess the importance of each concept. To this end, we propose a novel Composite Vision-Language Score $C_S$ - a weighted score, measuring the relevance of a concept node to both the image and the root of the concept tree. $C_S$ for a node $n^l$ is composed of a Visual Relevance Score $V_S$, which measures the probability of a concept being present in the image, and a Linguistic Relevance Score $L_S,$ which measures the probability of a concept entailed by the caption. Mathematically,
\begin{equation}
    C_S(n^l) = \alpha L_S(n^l,C) + (1-\alpha) V_S(I,n^l)
    \label{eq:alpha}
\end{equation}
where $\alpha$ is a tunable hyperparameter assigning weight to the linguistic relevance of the concept node. Intuitively, the Composite Score balances the over-representation of a concept in the linguistic bias with the grounding in the vision context. This ensures that often co-existing concepts not occurring in an image will be scored relatively lower than factors that are rare but actually grounded in the image. We will discuss implementation details of $V_S$ and $L_S$ in Section~\ref{sec:exp}.

\subsection{Dynamic Path Selection}
\noindent\textbf{Greedy Path Exploration w/ Beam Search.}
Once the concept tree is constructed, the final prediction score is calculated by \textit{path finding} in the tree. Note that a particular path $p$ through the tree is grounded with the morphological entity $e\in E$ as the starting node. A path through the concept tree rooted at $e$ thus is of the form $p = \{e, n^1,n^2,..n^l \} \in P $ where $P$ is the set of all possible paths. The composite scores are computed for each path as $W_p = \{C_S(n),~n\in p\}$. As can be observed, multiple different paths can be explored in the concept tree. To find the ideal reasoning pathway, we implement a function $SRCH$ to search for the ideal reasoning path weight $\hat{W_p}$ in two variants: (1) Greedy: Select the next node based on the highest composite score of each child node ($SRCH_{max}$), and (2) Beam: Consider multiple paths together ($SRCH_{Beam}$) by selecting $k$ maximum composite score nodes and subsequently considering the one with maximum path weight. The ideal $\hat{W_p}$ is treated as the System-2 output of the concept tree, as shown in Figure~\ref{fig:schematic}(c). The final output is computed as:
\begin{equation}
     \beta*f(I,C) +  (1-\beta)*\hat{W_p}
     \label{eq:beta}
\end{equation}
The hyperparameter $\beta$ controls the influence of the System-2 reasoner on VLM outputs.

\noindent\textbf{Neurosymbolic Reasoning along a Path.}
Finally, we discuss the interpretable reasoning along a path for improved interpretability. We consider two logical operations - $AND$ ($\land$) and $OR$ ($\lor$) to combine nodes along a path. Each node at a level $n^l$ and its corresponding expansion on the next level $n^{l+1}$ can be treated as either relating to each other's occurrence, i.e., $n^l \land n^{l+1}$ or independent, i.e., $n^l \lor n^{l+1}$. We consider both scenarios in interpretability analysis. In the result section, we demonstrate how conjoined reasoning affects System-2 interpretability where nodes chained together form a neurosymbolic rule for interpretability.


\section{Experiments}
\label{sec:exp}
\subsection{Datasets and Model Descriptions}
\noindent \textbf{Evaluation Datasets:} We utilize four benchmark compositionality datasets to validate our methodology. The datasets WinoGround \cite{thrush2022winoground}, EqBench \cite{wang2023equivariant}, and ColorSwap \cite{burapacheep2024colorswap} study the compositionality relationships between object semantics, positions, and colors, respectively. SugarCrepe \cite{hsieh2023sugarcrepe} benchmark is more fine-grained with both attributes and semantic relationships interlinked. As described in Section~\ref{sec:problem-setting}, each sample in the first three datasets consists of two images $I_0,I_1$ and two corresponding captions $C_0,C_1$. The objective remains to solve the \textit{Text, Image, and Group} tasks. For SugarCrepe, the task is simplified only to Text Score, as only a single image is provided. As EqBench has more than 10k samples, we sample a 2500-sample test set for all our experiments.

\noindent \textbf{Validation Data:} As Winoground, ColorSwap, and EqBench share a similar structure, we utilize 1000 random samples from the EqBench (Light) subset for hyperparameter tuning (disjoint from test). For ColorSwap, we utilize its train set for hyperparameter tuning. 

\noindent \textbf{Models:} We consider seven open-sourced VLMs of varying sizes from four different model families. InstructBLIP-XXL \cite{dai2023instructblip}, LLava \cite{liu2023llava}, Qwen \cite{bai2023qwen} and InternVL \cite{chen2024internvl}. We consider two model settings - models with parameter counts of 8 billion parameters (LLaVA-1.5-7b, LLaVA-1.6-7b, Qwen-7b, InternVL-8b) and 13 billion parameter range (InstructBLIP-XXL, LLaVA-1.5, LLaVA-1.6).

\noindent \textbf{Evaluation Metric:} We utilize VQAScore \cite{lin2025evaluating} as the raw VLM-only baseline. VQAScore treats VLM outputs as discriminative rather than generative by comparing token probabilities. We compare VQAScore as the baseline with Caption Score which represents an exact match of generated tokens while VisualGPTScore \cite{lin2023visualgptscore} considers the exact match of the generated sequence. We compare the methods in Table~\ref{tab:eval-metrics}. We observe VQAScore to be the best metric. 

\begin{table}[h]
\centering
\renewcommand{\arraystretch}{1.0}
\resizebox{0.47\textwidth}{!}{
\begin{tabular}{ccc}
\toprule
\textbf{Item} & \textbf{WinoGround} & \textbf{EqBench} \\
\midrule
Caption Score  & 17.50  &  16.75   \\
VisualGPTScore \cite{lin2023visualgptscore}   & 27.50  & 20.25  \\
VQA Score \cite{lin2025evaluating}    & 29.25    & 21.75   \\
\bottomrule
\end{tabular}}
\caption{\small Group Score comparisons of various evaluation baselines on the LLaVA-1.5-7B model. We utilize VQAScore as a metric due to its discriminative nature which alleviates the limited language generation capabilities of VLMs.}
\label{tab:eval-metrics}
\end{table}


\subsection{Implementation Details}
\noindent \textbf{LLM Reasoner Settings.} We utilize instruction-tuned LLama-3.1-8b with a total of 8 billion parameters as the LLM reasoner. The LLM reasoner is comparable to the size of each baseline VLM. For all steps, we utilize a temperature setting of 0 due to the deterministic nature of the task. We utilize $M=2$, i.e., the number of morphological entities and a splitting factor value $S=3$ with depth $L=3$. Note that larger LLMs can be utilized for generating the concept set, however, our study focuses on the `Multiple' setting (Table~\ref{tab:comparision} with constraints on the size of the LLM being close to the VLMs.

\noindent \textbf{Hyperparameter Settings.} We tune all hyperparameters on the validation set. For the Winoground and EqBench datasets, we select the values of $\alpha=0.6$, and $\beta=0.8$. For the ColorSwap dataset,  we select the values of $\alpha=0.5$ while the value of $\beta=0.8$. 

\noindent \textbf{Visual and Linguistic Score Computation.}
We calculate Visual Relevance Score $V_S$ as:
\begin{equation}
    V_S(I,C) = \mathbb{P}_{VLM}(\text{``yes''} ~ |~ I,~ C) 
\end{equation}
where $\mathbb{P}$ is the probability of predicting token ``yes'' by a VLM representing the presence of $C$ in image $I$ \cite{lin2025evaluating, hessel2021clipscore}.

Similarly, to quantify linguistic relations, we propose the Linguistic Relevance Score $L_S$. We calculate $L_S$ between textual inputs $C_1$ and $C_2$ using an LLM as:
\begin{equation}
         L_S(C_1,C_2) = \mathbb{P}_{LLM}(\text{``yes''} ~ |~C1,C2)
\end{equation}
where $\mathbb{P}$ is the probability of perfect non-contradiction.

\subsection{Comparison Baselines}
For reproducibility, we restrict baselines to methods that are runnable with fully open-source models. We replicate and compare COCO-Tree against the best-performing baseline in Single Setting - CCoT \cite{mitra2024compositional}. For `Multiple' setting, none of the baselines compare against CCoT and VQAScore as evidenced in \cite{cascante-bonilla2025natural} (Refer Appendix for more details). For a fair comparison, we compare CECE in a realistic setting - with base VLM and equally sized LLM to COCO-Tree in Table~\ref{tab:comparision-baseline}. We utilize LLaVA-1.5-7B as the VLM and LLaMA-3.1-8B as the LLM. We report the performance on 200 samples sampled from the WinoGround and EqBench datasets randomly. We point out that CECE's original setting is not comparable to our approach, as shown in Table~\ref{tab:comparision}.

\begin{table}[h]
\centering
\renewcommand{\arraystretch}{1.0} 
\resizebox{0.45\textwidth}{!}{
\begin{tabular}{ccc}
\toprule
\textbf{Method} & \textbf{WinoGround} & \textbf{EqBench}  \\
\midrule
VQAScore \cite{lin2025evaluating} & 29.00 & 24.50 \\
CECE \cite{cascante-bonilla2025natural} & 32.50  & 34.25 \\
COCO-Tree (Ours) & \textbf{35.00} & \textbf{37.50} \\
\bottomrule
\end{tabular}}
\caption{\small Group Scores for VQAScore, CECE and COCO-Tree on Llava-1.5-7B with LLaMA-3.1-8B on the Winoground and EqBench subsets based on our replication. Note: We utilize our own replication for all methods with hyperparameters taken from original papers.}
\label{tab:comparision-baseline}
\end{table}

\begin{table*}[t]
\centering
\setlength{\tabcolsep}{6pt}
\renewcommand{\arraystretch}{1.05}
\resizebox{\textwidth}{!}{%
\begin{tabular}{c|c|ccc|ccc|ccc|c}
\toprule
\textbf{Model} & \textbf{Method} & \multicolumn{3}{c|}{\textbf{WinoGround}} &
\multicolumn{3}{c|}{\textbf{EqBench}} & \multicolumn{3}{c|}{\textbf{ColorSwap}} &
\multicolumn{1}{|c}{\textbf{SugarCrepe}} \\
\midrule
& & \textbf{Text} & \textbf{Image} & \textbf{Group} &
\textbf{Text} & \textbf{Image} & \textbf{Group} &
\textbf{Text} & \textbf{Image} & \textbf{Group} & \textbf{Mean} \\
\midrule
\multirow{4}{*}{\textbf{LLaVA-1.5-7b}}
& \textbf{VQAScore} & 44.50 & 43.75 & 29.25 & 34.75 & 42.00 & 21.75 & 89.33 & 77.33 & 74.00 & 87.36 \\
& \textbf{CCoT} & 44.25 & 41.25 & 26.50 & 35.50 & 42.00 & 21.75 & 88.33 & 76.66 & 72.00 & 88.55 \\
& \textbf{COCO-Tree (Max)} & \textbf{48.25} & \textbf{44.50} & \textbf{35.00} & \textbf{39.50} & \textbf{42.00} & \textbf{28.25} & \textbf{93.33} & \textbf{91.00} & \textbf{87.66} & 89.84 \\
& \textbf{COCO-Tree (Beam)} & \textbf{48.75} & \textbf{46.50} & \textbf{35.25} & \textbf{43.25} & \textbf{43.75} & \textbf{37.50} & \textbf{93.33} & \textbf{91.00} & \textbf{87.66} & \textbf{90.67} \\
\midrule
\multirow{4}{*}{\textbf{LLaVA-1.6-7b}}
& \textbf{VQAScore} & 51.50 & 52.00 & 36.50 & 21.75 & 48.00 & 18.00 & 87.63 & 82.27 & 76.92 & 88.90 \\
& \textbf{CCoT} & 52.00 & 52.50 & 37.50 & 28.00 & 48.25 & 21.50 & 87.63 & 82.27 & 76.92 & 89.28 \\
& \textbf{COCO-Tree (Max)} & \textbf{56.00} & \textbf{52.00} & \textbf{40.50} & 42.00 & \textbf{49.25} & 28.00 & \textbf{93.33} & \textbf{91.00} & \textbf{87.66} & \textbf{90.28} \\
& \textbf{COCO-Tree (Beam)} & \textbf{56.00} & \textbf{52.00} & \textbf{40.50} & \textbf{42.25} & 49.00 & \textbf{37.25} & \textbf{93.33} & \textbf{91.00} & \textbf{87.66} & \textbf{90.28} \\
\midrule
\multirow{4}{*}{\textbf{Qwen-7b}}
& \textbf{VQAScore} & 59.00 & 54.00 & 45.50 & 35.25 & 45.75 & 24.75 & 92.00 & 90.67 & 87.33 & 85.22 \\
& \textbf{CCoT} & 59.00 & 54.00 & 45.50 & 35.25 & 45.75 & 24.75 & 92.00 & 90.67 & 87.33 & 85.67 \\
& \textbf{COCO-Tree (Max)} & 59.50 & \textbf{58.50} & 45.50 & 41.00 & \textbf{46.00} & 30.50 & \textbf{93.33} & \textbf{91.33} & \textbf{88.67} & 85.67 \\
& \textbf{COCO-Tree (Beam)} & \textbf{60.50} & \textbf{58.50} & \textbf{47.00} & \textbf{42.75} & 44.25 & \textbf{36.25} & \textbf{93.33} & 91.00 & 87.66 & \textbf{86.25} \\
\midrule
\multirow{4}{*}{\textbf{InternVL-8b}}
& \textbf{VQAScore} & 64.00 & 62.00 & 51.25 & 43.50 & 57.75 & 37.50 & 93.33 & 91.33 & 88.67 & 94.52 \\
& \textbf{CCoT} & 64.00 & 62.00 & 51.25 & 43.50 & 57.75 & 37.50 & 93.33 & 91.33 & 88.67 & 94.52 \\
& \textbf{COCO-Tree (Max)} & \textbf{64.50} & \textbf{62.50} & \textbf{52.50} & \textbf{45.00} & \textbf{57.75} & \textbf{39.50} & \textbf{96.00} & \textbf{94.67} & \textbf{92.67} & \textbf{95.25} \\
& \textbf{COCO-Tree (Beam)} & \textbf{64.00} & \textbf{62.50} & \textbf{52.50} & \textbf{45.00} & \textbf{57.75} & \textbf{39.50} & \textbf{96.00} & \textbf{94.67} & \textbf{92.67} & \textbf{95.25} \\
\midrule
\multirow{4}{*}{\textbf{InstructBLIP-XXL}}
& \textbf{VQAScore} & 41.50 & 41.25 & 27.75 & 21.75 & 48.00 & 18.00 & 88.00 & 88.67 & 83.33 & 90.36 \\
& \textbf{CCoT} & 40.50 & 40.25 & 25.00 & 26.50 & 48.00 & 19.50 & 84.55 & 87.00 & 82.15 & 89.74 \\
& \textbf{COCO-Tree (Max)} & \textbf{48.00} & \textbf{47.75} & \textbf{38.75} & \textbf{29.75} & \textbf{52.50} & \textbf{26.75} & \textbf{89.33} & \textbf{89.33} & \textbf{84.00} & \textbf{90.67} \\
& \textbf{COCO-Tree (Beam)} & \textbf{48.00} & \textbf{47.75} & \textbf{38.75} & \textbf{29.75} & \textbf{52.50} & \textbf{26.75} & \textbf{89.33} & \textbf{89.33} & \textbf{84.00} & \textbf{90.67} \\
\midrule
\multirow{4}{*}{\textbf{LLaVA-1.5}}
& \textbf{VQAScore} & 45.00 & 48.00 & 31.50 & 35.50 & 41.50 & 21.00 & 91.33 & 91.67 & 88.33 & 89.28 \\
& \textbf{CCoT} & 46.50 & 47.50 & 32.50 & 34.50 & 40.00 & 20.00 & 91.33 & 91.67 & 88.33 & 89.42 \\
& \textbf{COCO-Tree (Max)} & \textbf{55.00} & 46.50 & 40.50 & \textbf{45.00} & \textbf{42.50} & \textbf{28.50} & \textbf{94.33} & \textbf{92.33} & \textbf{88.67} & \textbf{90.24} \\
& \textbf{COCO-Tree (Beam)} & 54.00 & \textbf{53.00} & \textbf{42.00} & \textbf{45.00} & \textbf{42.50} & \textbf{28.50} & \textbf{94.33} & \textbf{92.33} & \textbf{88.67} & \textbf{90.24} \\
\midrule
\multirow{4}{*}{\textbf{LLaVA-1.6}}
& \textbf{VQAScore} & 45.00 & 48.00 & 31.50 & 35.50 & 41.50 & 21.50 & 92.00 & 90.67 & 87.33 & 91.71 \\
& \textbf{CCoT} & 45.50 & 48.00 & 32.00 & 35.50 & 42.00 & 22.00 & 91.33 & 90.67 & 86.67 & 92.28 \\
& \textbf{COCO-Tree (Max)} & \textbf{59.00} & \textbf{54.00} & \textbf{49.50} & \textbf{43.00} & \textbf{42.50} & \textbf{25.50} & \textbf{96.00} & \textbf{94.67} & \textbf{92.67} & \textbf{93.85} \\
& \textbf{COCO-Tree (Beam)} & \textbf{59.00} & \textbf{54.00} & \textbf{49.50} & \textbf{43.00} & \textbf{42.50} & \textbf{25.50} & \textbf{96.00} & \textbf{94.67} & \textbf{92.67} & \textbf{93.85} \\
\bottomrule
\end{tabular}}
\caption{\small Compositionality performance of COCO-TREE (ours) compared to VQAScore \cite{lin2025evaluating} and CCoT \cite{mitra2024compositional}. “Max” and “Beam” denote the path selection strategy. Evaluated for 7 VLMs on 4 datasets: WinoGround, EqBench, ColorSwap, and SugarCrepe.}
\label{tab:main-results}
\end{table*}

\subsection{Results and Analysis}
\noindent \textbf{Quantitative Compositionality Performance: } We report the performance on the selected baselines and VLMs in Table~\ref{tab:main-results}. Note that we compare the replicated baselines VQAScore and CCoT. In certain cases, CCoT degrades performance compared to VQAScore, implying that the context generated (scene graph) during the first-stage prompting is inaccurate and detrimental to effective compositionality understanding. We report the performance of COCO-Tree using the Max and Beam path selection strategies as discussed in Section~\ref{sec:method}:
\begin{itemize}[leftmargin=*, parsep=0pt, itemsep=0pt, topsep=0pt]
\item \textbf{Winoground:} For the LLaVA family of models (Rows-1,2,6,7) and InstructBLIP-XXL (Row-5), we observe COCO-Tree outperforms all baselines by an average of 5\%. For the Qwen and InternVL models, we observe performance gains on all settings by an average of 2\%.
\item \textbf{EqBench:} We observe a consistent improvement over all baselines across all models by an average of 5-8\% - a substantial improvement, demonstrating the efficacy of our approach. 
\item \textbf{ColorSwap:} Owing to the reduced complexity of ColorSwap as compared to WinoGround and EqBench, most models already achieve high performance. COCO-Tree successfully improves performance across all models, with an average of 4-6\%, a significant improvement on already high numbers. 
\item \textbf{SugarCrepe:} The performance on SugarCrepe benchmark is reported as the mean value of Text Score over all sub-sets (Refer to Appendix for detailed analysis). We observe that COCO-Tree outperforms all baselines by about 2\%. Note that the performance on SugarCrepe is already extremely high.
\end{itemize}
One peculiar result observed is the modest gains on the image scores as compared to text and group scores for COCO-Tree as compared to baselines. This observation indicates that improved linguistic reasoning alone cannot significantly improve the Image selection task.

\noindent \textbf{Detailed Performance on sub-sets: } Next, we report the results of the best-performing configurations of COCO-Tree on the labeled subsets of the Winoground dataset and SugarCrepe datasets in the Appendix. We report the \textbf{win rate} of COCO-Tree over baselines, i.e. subsets where performance improves. We observe a 100\% win rate on WinoGround and an 89\% win rate on SugarCrepe.

\noindent \textbf{Computation Cost Analysis.} We compare the computation cost of our method with approaches in the `Multiple' model setting. Note that we require access to only a small-scale LLM reasoner (similar in scale to VLM). As compared to DSG which uses a 3-stage process (Prompt, Tuple, Question) our method uses only a 2-step process (tree construction and scoring). As compared to CECE, which requires inferences with a 70 billion parameter LLM, COCO-Tree uses an 8 billion parameter LLM. The upper bound of time complexity is directly dependent on the number of concept nodes - $\mathcal{O}$($M * S* L$).

\noindent \textbf{Discussion around variable performance improvements across datasets:} We attribute this behavior to Dataset Complexity - Winoground and Eqbench datasets require models to differentiate between multiple highly semantic abstract entities, where decomposing and concept-tree modeling help the most in improving understanding, which is reflected in performance gains (5\% gains on Winoground and 
10\% on EqBench). However, on datasets like SugarCrepe, where the images differ only on one single attribute, constructing a vast concept tree yields relatively modest gains (1-2\%) on a very high off-the-shelf baseline performance. This, in turn, implies that modern VLMs are good at understanding compositionality in simpler cases but less effective when image complexity increases. The gains on complex datasets are a testament to our approach. Additionally, newer state-of-the-art VLMs (InternVL, Qwen) are pre-trained on broader, more holistic datasets and can capture and internalize semantic priors more effectively. On these models, COCO-Tree still provides significant refinement (2-5\%). The performance jump on older VLMs remains very high (10\%) across multiple datasets.



\subsection{Ablation Studies}
\noindent We utilize the Winoground dataset for ablation analysis due to its challenging nature and smaller size and test on two significantly different models - LLaVA-1.5-7b and InstructBLIP-XXL.

\noindent \textbf{Tree Hyperparameters.} We report extensive ablations around hyperparameters affecting the concept-tree structure in Table~\ref{tab:ablation-hparams} as compared to baseline VQAScore.  First, we report results of varying depths of the concept tree exploration in the fourth column, for tree depths from 1 to 3. We observe that exploring the concept tree to a deeper depth increases the quality of concepts generated and hence, improves performance across all VLMs. In the next 3 columns, we explore the effect of the splitting factor $S$, which controls the number of children nodes for each node in the tree.

\begin{table}[h]
    \centering
    \setlength{\tabcolsep}{6pt}
\renewcommand{\arraystretch}{1.05}
  \resizebox{\linewidth}{!}{%
    \begin{tabular}{l|c|c|ccc|ccc|cc}
      \toprule
      \textbf{Model} & \textbf{Task} & \textbf{Base} &
        \multicolumn{3}{c|}{\bm{$L$}} &
        \multicolumn{3}{c|}{\bm{$S$}} &
        \multicolumn{2}{c}{\bm{$M$}} \\ \cline{4-11}
      & & & 1 & 2 & 3 & 2 & 3 & 4 & 2 & 4 \\ \midrule
      \multirow{3}{*}{\textbf{L-1.5}}
        & Text  & 44.50 & 46.00 & 46.75 & 48.25 & 47.75 & 48.25 & 48.25 & 48.25 & 48.50 \\
        & Image & 43.75 & 42.00 & 43.50 & 44.50 & 43.25 & 44.50 & 44.50 & 44.50 & 44.75 \\
        & Group & 29.25 & 33.25 & 34.25 & 35.00 & 34.00 & 35.00 & 35.00 & 35.00 & 35.25 \\ \midrule
      \multirow{3}{*}{\textbf{I-BLIP}}
        & Text  & 41.50 & 39.50 & 44.25 & 48.00 & 40.50 & 48.00 & 48.00 & 48.00 & 48.50 \\
        & Image & 41.25 & 42.50 & 45.25 & 47.75 & 45.00 & 47.75 & 47.75 & 47.75 & 48.00 \\
        & Group & 27.75 & 34.50 & 35.75 & 38.75 & 34.00 & 38.75 & 38.75 & 38.75 & 39.00 \\ 
        \bottomrule
    \end{tabular}}
    \caption{\small Effect of depth, split factor \(S\), and caption splits \(M\) on Winoground. We evaluate on two models - LLaVA-1.5-7b (L-1.5) and InstructBLIP-XXL (I-BLIP).}
    \label{tab:ablation-hparams}
\end{table}

We observe that increasing $S$ improves performance before plateauing. Next, we report the effect of $M$ representing number of morphological entities the caption is split into. We observe that increasing $M$ improves performance, but also increases size of the tree by two orders of magnitude.

\begin{figure}
    \centering
    \includegraphics[width=0.48\textwidth]{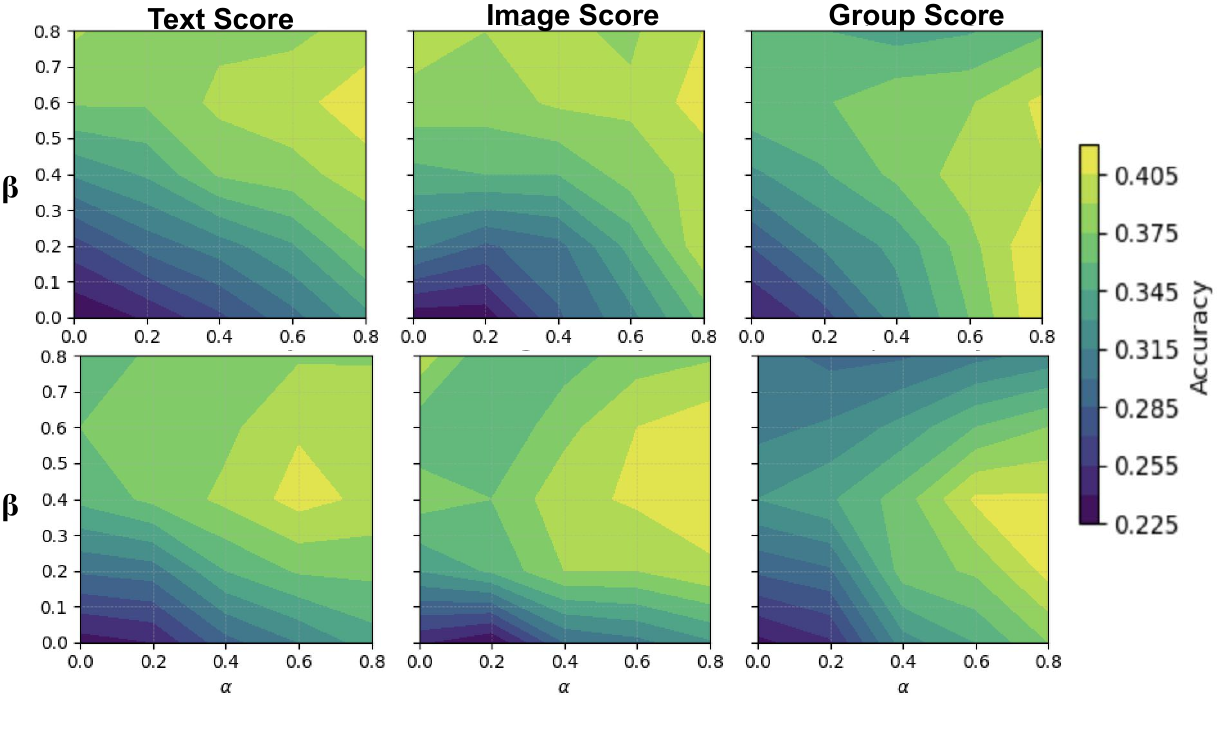}
    \caption{\small Ablation study on the impact of composite score hyperparameters $\alpha$ and $\beta$. The color gradient represents the accuracy with deep Yellow being the maximum and deep Purple being the minimum scores. Top: LLaVA-1.5-7b, Bottom: InstructBLIP-XXL.}
    \label{fig:alpha-beta-ablation}
\end{figure}

\noindent \textbf{Composite Score Hyperparameters.} Next, we report the effect of $\alpha$, which controls the effect of linguistic relevance and $\beta$ which controls the effect of System-2 reasoning on the final output in Figure~\ref{fig:alpha-beta-ablation} for models - LlaVA-1.5-7b and InstructBLIP-XXL. We observe that too low or too high of $\beta$ degrades performance, which is understandable as purely System-1 or System-2 outputs do not capture the fine-grained semantics for compositionality. We also observe that intermediate values of $\alpha$ produce the highest performance as both linguistic and visual references contribute equally to the prediction.

\subsection{Interpretability Analysis}
Lastly, we highlight that the selected concept pathways in our approach can be utilized to form a neurosymbolic rule. We utilize a large multimodal model (GPT-4o) as the `judge' to assign entailment scores given a constructed rule. For example, the highest scoring rule in Figure~\ref{fig:motivation} is $P$=\{`consuming a snake', 'snake in bird's mouth', 'a snake is being held by a bird'\}. The rules are constructed through the ${AND}$ and ${OR}$ operations and passed through to the judge and an entailment score is calculated. For instance, the $AND$ rule takes the form - ``consuming a snake $AND$ snake in bird's mouth $AND$ a snake is being held by a bird $\implies$ bird eats snake''. 

\begin{figure}[h]
    \centering
    \includegraphics[width=0.48\textwidth]{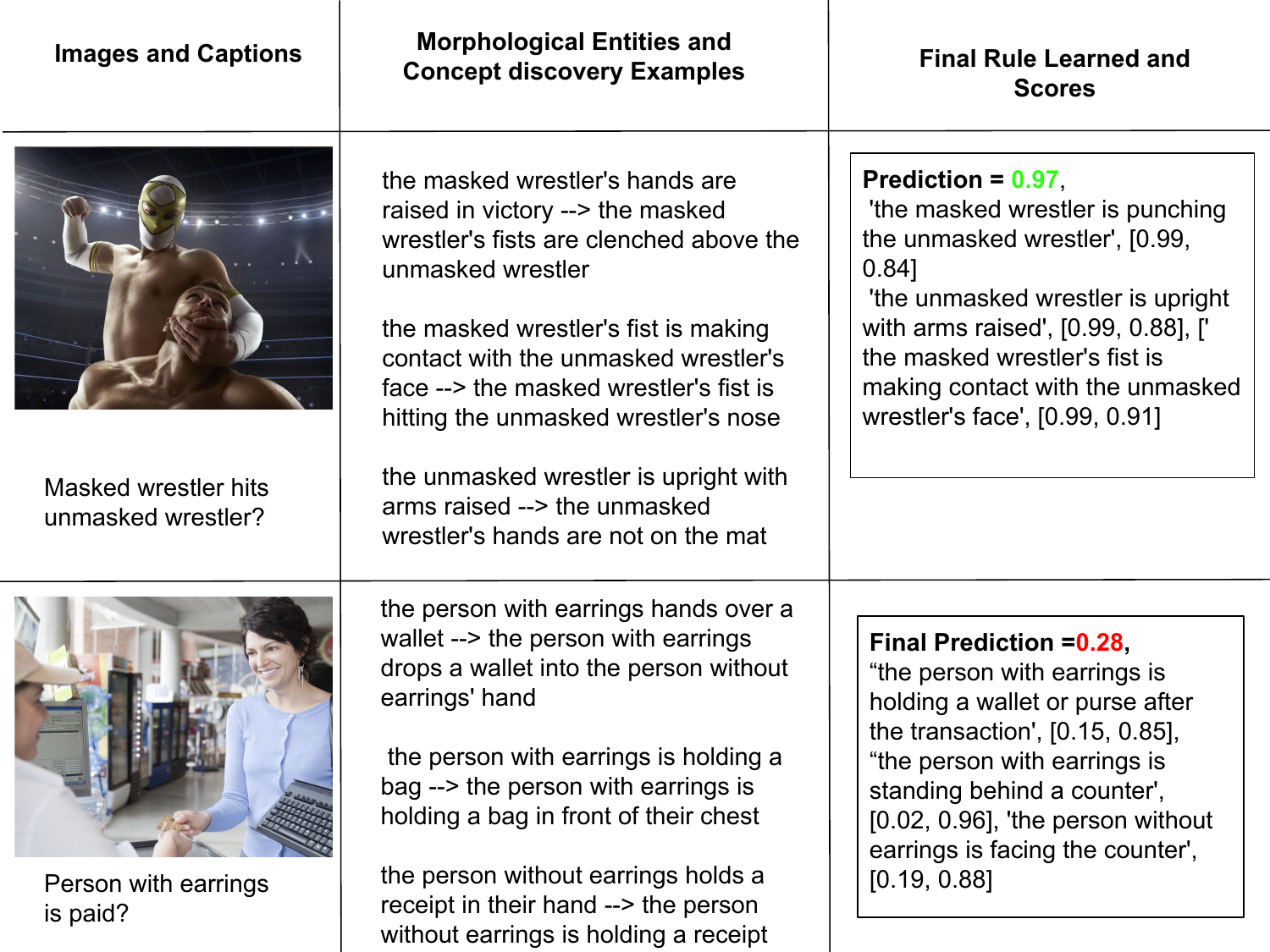}
    \caption{\small The reasoning pathway for two randomly chosen test samples from the Winoground dataset using LLava-1.5-7b. Prediction scores represent the reasoning path probability of a positive and a negative sample.}
    \label{fig:reasoning-path}
\end{figure}

We report the average entailment scores for the Winoground and EqBench datasets in Table~\ref{tab:entailment}. We consider 3 settings wherein only rules, only caption,  the rules, and caption are fed into the judge with and without an image. We observe that the rule generated using our approach gives higher entailment scores from the judge as compared to only feeding in the caption with $\lor$ rules giving higher confidence. This observation validates the interpretability of the reasoning paths in the concept tree. A visual description is shown in Figure~\ref{fig:reasoning-path}. We observe that both AND and OR rules are beneficial over just captions in both with and without image cases, with the OR rules showing better results. This validates the quality of the concept pathway and the constructed rule. We include more examples in the Appendix.

\begin{table}[ht]
\centering
\setlength{\tabcolsep}{3pt}
\renewcommand{\arraystretch}{0.9}
\resizebox{0.4\textwidth}{!}{
\begin{tabular}{l|cc|cc}
\toprule
 & \multicolumn{2}{c|}{\textbf{Winoground}} & \multicolumn{2}{c}{\textbf{EqBench}} \\ 
\midrule
 & \textbf{w/o Image} & \textbf{w/ Image} & \textbf{w/o Image} & \textbf{w/ Image} \\ 
\midrule
\textbf{Only $\land$ Rule} & 0.74 & 0.85 & 0.79 & 0.96 \\
\textbf{Only $\lor$ Rule} & 0.81 & 0.88 & 0.85 & 0.96 \\  
\textbf{Only Caption} & 0.65 & 0.92 & 0.83 & 0.96\\  
\textbf{$\land$ Rule + Caption} & 0.91  & 0.98 & 0.98 & 0.99 \\  
\textbf{$\lor$ Rule + Caption} & \textbf{0.93} & \textbf{0.98} & \textbf{0.98} & \textbf{0.99} \\ 
\bottomrule
\end{tabular}}
\caption{We show the effect of rules, captions and a combination of both in the confidence of entailment for the Winoground and EqBench datasets.}
\label{tab:entailment}
\end{table}

\noindent \textbf{Sanity Checking LLM outputs:} To check whether the LLM produces the required concepts in the tree, we manually analyze a handful of examples as detailed in Figures 10 and 11 in the Appendix. In Figure 11, the concepts generated are: `the object is made of absorbent material', the towel is not on the ground', ‘the person's arms are not moving
in a throwing motion’ which are all correct concepts for predicting the caption question - `Person is holding a towel?'.

\noindent \textbf{Possible Failure Cases:} As in all LLM and VLM studies, there is a distinct chance of some concepts being hallucinated, which miss our heuristic filtering process. On the other hand, there is also a possibility of VLM incorrectly attributing the presence and/or absence of a concept in an image with high certainty. Due to the Composite Vision-Language Score, the LLM and VLM scores are balanced. We also discuss some cases around the hallucination of the VLM and LLM components individually, and how COCO-Tree handles these cases in the Appendix.
\section{Conclusion}
\label{sec:conclusion}
In this work, we introduced  (COCO-Tree), a novel framework that augments low-resource VLMs by incorporating neuro-symbolic concept trees derived from LLMs. Our approach not only improves compositional generalization but also provides interpretability by offering explicit rationales for model predictions. Empirical evaluations on Winoground, EqBench, ColorSwap, and SugarCrepe benchmarks demonstrate that COCO-Tree significantly enhances compositional reasoning, yielding an average improvement of 10\% across multiple open-source VLMs. These findings highlight the potential of synergizing VLMs with LLMs to overcome compositional limitations.

\section*{Acknowledgment}
This work is supported in part by the US National Science Foundation (NSF) and the National Institute of Health (NIH) under grants IIS-2106913, IIS-2538206, IIS-2529378, CCF-2217071, CNS-2213700, and R01LM014012-01A1. Any recommendations expressed in this material are those of the authors and do not necessarily reflect the views of NIH or NSF.

\section*{Limitations}
(1) Hallucinations: As with all LLM-related research work, COCO-Tree can suffer from ill effects of hallucination. As each node in the concept tree is generated by a frozen LLM, spurious or irrelevant concepts can be introduced and subsequently weighted into the final score, occasionally degrading accuracy or producing brittle, self-reinforcing failure cases. (2) Resource utilization: although the 8B LLM is modest by today’s standards, maintaining multiple VLM feature maps plus a breadth-concept tree in memory scales exponentially, and limits edge deployment. 
(3) Inference-time complexity: the joint search explores textual branches and runs a forward pass on the VLM for each candidate, yielding relatively large compute utilization which can be impractical for some users. Future work will explore refining COCO-Tree’s neuro-symbolic structures to further boost compositional understanding and extend its applicability to broader vision-language tasks. (4) Compositionality datasets are currently designed primarily to identify relations between two entities. Even though our method is generalizable to an arbitrary number of entities, it is possible that it can underperform.

\bibliography{custom}

\appendix
\clearpage
\section*{Appendix}
\label{sec:appendix}

\section{Detailed Dataset Description}
\noindent \textbf{Winoground} \cite{thrush2022winoground} consists of 400 data points, each data point consisting of 2 semantically opposite images sampled from Getty Images. The captions are structurally similar but semantically opposite (as shown in Figure~\ref{fig:motivation}. In addition, each data point also consists of a tag that describes the high-level description of the type of captions and/or attributes in the image. These are - Object, Relation, and Both. For some data points, an additional tag labeled as `Symbolic' and `Pragmatic' is also present which signifies whether the content in the caption can be directly seen in the image (Symbolic) or not (Pragmatic).

\noindent \textbf{EqBench} \cite{wang2023equivariant}: The EqBench dataset consists of data points with pairs of images with `minimal
visual semantic changes'. The images are sampled from both a video library and a synthetic image generator engine covering diverse image domains. The minimal changes in the images include pairs with accurate semantic changes in action, location, and attributions.

\noindent \textbf{ColorSwap} \cite{burapacheep2024colorswap} The ColorSwap dataset consists of
1000 Winoground-style quadruplets focused on colour-to-object binding, built from 2\,000 diffusion-generated images with human verification. Each caption pair is lexically identical except for swapped color adjectives, creating minimal word-order contrasts. Provides 2000 positives and 2000 color-swapped hard negatives. It benchmarks whether models can correctly ground colors to objects and respect word order.

\noindent \textbf{SugarCrepe} \cite{hsieh2023sugarcrepe}: The SugarCrepe dataset consists of 7512 COCO-2017–derived image–caption examples; each gives one positive caption and an LLM-generated, fluent hard negative. Hard negatives span seven edit categories (REPLACE-OBJ/ATT/REL, SWAP-OBJ/ATT, ADD-OBJ/ATT), probing fine-grained compositionality. Adversarial refinement eliminates annotation artifacts, driving blind text models to random-chance (50\%) accuracy.

\section{Detailed Model Descriptions}
\begin{itemize}
    \item \textbf{InstructBLIP-Flan-T5-XXL} \cite{dai2023instructblip} is a multimodal AI model designed for vision-language tasks, integrating the BLIP-2 (Bootstrapped Language-Image Pretraining) \cite{li2023blip} framework with Flan-T5-XXL, a powerful text-to-text transformer from Google's Flan-T5 series \cite{chung2022scaling}.

    \item \textbf{Llava-1.5} \cite{liu2024improved} is an advanced vision-language model (VLM) that integrates LLaMA \cite{dubey2024llama}  with a visual encoder for multimodal understanding. It uses improved visual encoders based on CLIP \cite{radford2021learning} and instruction tuning to generate more context-aware and detailed responses. We utilize the 7 billion parameter version for all experiments.
    
    \item \textbf{Llava-1.6 (Llava-Next)} \cite{llava2024} is the next iteration of the LLava-1.5 family of models utilizing a stronger image encoder and diverse multimodal training data. Llava-Next uses an instruction-tuned LLM framework built on Llama models. In our experiments, we utilize the 7 billion version built on top of the Vicuna \cite{vicuna2023} model.

    \item \textbf{Qwen2-VL-7B} \cite{yang2024qwen2} extends the Qwen-2 language family with a dual-resolution vision encoder and gated cross-modal fusion. Multi-stage instruction tuning—captioning, VQA, and conversational steps—equips the 7 billion parameter model with strong zero-shot reasoning and grounding capabilities while keeping GPU memory use modest.

    \item \textbf{InternVL-2.5-8B} \cite{chen2024internvl} combines an 8 billion parameter InternLM-2.5 backbone with a ViT-G/14 image encoder and dense cross-attention bridges. Progressive contrastive pre-training followed by instruction tuning yields state-of-the-art performance on captioning, grounding, and region-level understanding tasks.
\end{itemize}

\begin{figure}[t]
\begin{tcolorbox}[colback=grey,colframe=black,title=Prompt template for $F_{SMD}$]
You are a helpful chatbot. Divide the caption into {M} smaller independent statements which entail the caption based on Subject and Object. Caption: \{$C$\}. The output format is:\\ 1. Subject 2. Object \\ Assistant:
\end{tcolorbox}
\caption{Prompt template used to generate morphological entities for function $F_{SMD}$ using an LLM.}
\label{fig:prompt-morpho}
\begin{tcolorbox}[colback=grey,colframe=black,title=Prompt template for $F_{RCE}$]
You are a helpful chatbot. List \{S\} binary visual concepts to verify the \{$n^l_i$\}. Ensure the outputs are possible for \{$C$\}. Answer in small phrases and focus on verifiable things like objects, locations, actions, etc. Output format is: 1. xxx 2. xxx 3. xxx 4. xxx 5. xxx.\\ Assistant: 
\end{tcolorbox}
\caption{Prompt template used to discover concepts for function $F_{RCE}$ using an LLM.}
\label{fig:prompt-concept}
\begin{tcolorbox}[colback=grey,colframe=black,title=Prompt template for $V_S$]
"<image> \{I\} Does this figure show: {C}? Please answer Yes or No."
\end{tcolorbox}
\caption{Prompt template used to calculate Visual Score $V_{S}$ using a VLM.}
\label{fig:prompt-vs} 
\begin{tcolorbox}[colback=grey,colframe=black,title=Prompt template for $L_S$]
"Given we observe \{C1\}. Is it possible \{C2\}? Answer yes or no. Assistant: "
\end{tcolorbox}
\caption{Prompt template used to calculate Linguistic Score $L_{S}$ using an LLM.}
\label{fig:prompt-ls} 
\vspace{-10pt}
\end{figure}

\section{Prompt Templates}
To ensure reproducibility, we give the exact prompt templates for the concept tree construction functions $F_{SMD}$ in Figure~\ref{fig:prompt-morpho} and $F_{RCE}$ in Figure~\ref{fig:prompt-concept}. In addition, the prompt templates to compute $V_S$ are detailed in Figure~\ref{fig:prompt-vs} and $L_S$ is detailed in Figure~\ref{fig:prompt-ls}.


\section{Salient Details on Composite Score}
We compute the probability of a VLM for a caption $\mathcal{P}$ by transforming the caption into a binary question. Assume $\hat{F}_{VLM}[t]$ denote the \textit{output logits} for a VLM $F_{VLM}$ and a token $t$, the probability is calculated as:
\begin{equation}
    P_{VLM}(\text{``yes''}|I,C) = \frac{e^{\hat{F}_{VLM}[\text{``yes''}]}}{e^{\hat{F}_{VLM}[\text{``yes''}]} + e^{\hat{F}_{VLM}[\text{``no''}]}}
\end{equation}

Similarly, for an LLM calculating entailment between two phrases, assume $\hat{F}_{LLM}[t]$ denote the \textit{output logits} for a $F_{LLM}$ and a token $t$, the probability is calculated as:
\begin{equation}
    P_{LLM}(\text{``yes''}|C_1,C_2) = \frac{e^{\hat{F}_{LLM}[\text{``yes''}]}}{e^{\hat{F}_{LLM}[\text{``yes''}]} + e^{\hat{F}_{LLM}[\text{``no''}]}}
\end{equation}
The Composite Score calculations utilize the probability $P_{LLM}$ and $P_{VLM}$ as discussed.

\section{Robustness of COCO-Tree and Composite Score under Model Hallucinations}
COCO-Tree is robust enough to handle multiple error cases. We identify two sources of possible errors due to hallucinations: 1) the LLM generates concepts that are never encountered in the image, and 2) the LLM does not generate a prominent concept present in the image if the caption information is sparse. We solve both problems through the Composite Scoring mechanism, which balances the impact of both LLM and VLM scores on each node (Equation~\ref{eq:alpha}), and the Dynamic System-2 integration mechanism, which balances the impact of the concept tree on VLM outputs (Equation~\ref{eq:beta}). We provide more details below:

\textbf{Composite Scoring balances LLM and VLM scores}: Assume for an image, an ideal relevant concept is generated by the LLM, making its Linguistic Score = 0.99, while it is never found in the image, making its Visual Score = 0.01. If we only consider the VLM output, this concept would never be utilized for prediction. However, our Composite Score (assuming $\alpha=0.6$,$1-\alpha=0.4$) would bring the concept node's weight value to $0.99*0.6+0.4*0.01=0.594$, an intermediate value that would bring it into consideration of potential reasoning paths. With our proposed beam search approach, this concept has a high likelihood of being considered in one of the reasoning pathways (the typical concept node's composite score is between $0.5-0.8$, highlighting the effectiveness of composite scoring and COCO-Tree's beam search method. Note that the effective scoring depends on the aggregation of the scores for the entire path, so there is a negligible possibility of a completely hallucinated path.

\textbf{Dynamic System-2 integration balances concept-tree's output}: Let's consider the flip case, where the LLM completely misses a concept present prominently in the image, i.e., the concept node does not exist in the tree. As COCO-Tree aggregates the final output prediction as a weighted sum of VLM-only output and concept-tree output balanced by (Equation 8), such a concept would contribute heavily to VLM's prediction performance, improving prediction performance. Finally, as discussed in Section 4.6, our selected neurosymbolic pathway provides one possible explanation for a given sample.

In addition, we would also like to point out that hallucinations in generating concept-trees are also constrained during the node discovery procedure by grounding the generation to the root of the tree (i.e., the caption) (Refer Section 3.2) and ensuring each discovered concept node is entailed by the root node.

\section{Human Study to Test Concept Discovery}
We perform a 100-sample human study with 10 different humans, each of whom is given 10 samples sampled from each layer of the concept tree from the Winoground dataset. We ask every participant 2 questions for each concept - Q1: `Given \{concept node\}, do you think it may happen simultaneously with \{root node\} or not related at all?' and Q2: `Given \{concept node\}, do you think it is similar to \{parent node\} or not related at all?'. We report the \% Yes answers for both questions averaged across 10 participants:
\begin{table}[h]
    \centering
    \small
    \begin{tabular}{c|c}
        \toprule
         Root Node (Q1) & Parent Node (Q2)\\
         \midrule
         74\%& 79\% \\
         \bottomrule
    \end{tabular}
    \label{tab:human}
\end{table}
We observe that the non-relation (i.e. semantic drift) increases more from the root node than the parent node.

\section{Pseudocode for COCO-Tree}
\begin{figure*}[h]
    \centering
    \includegraphics[width=0.75\textwidth]{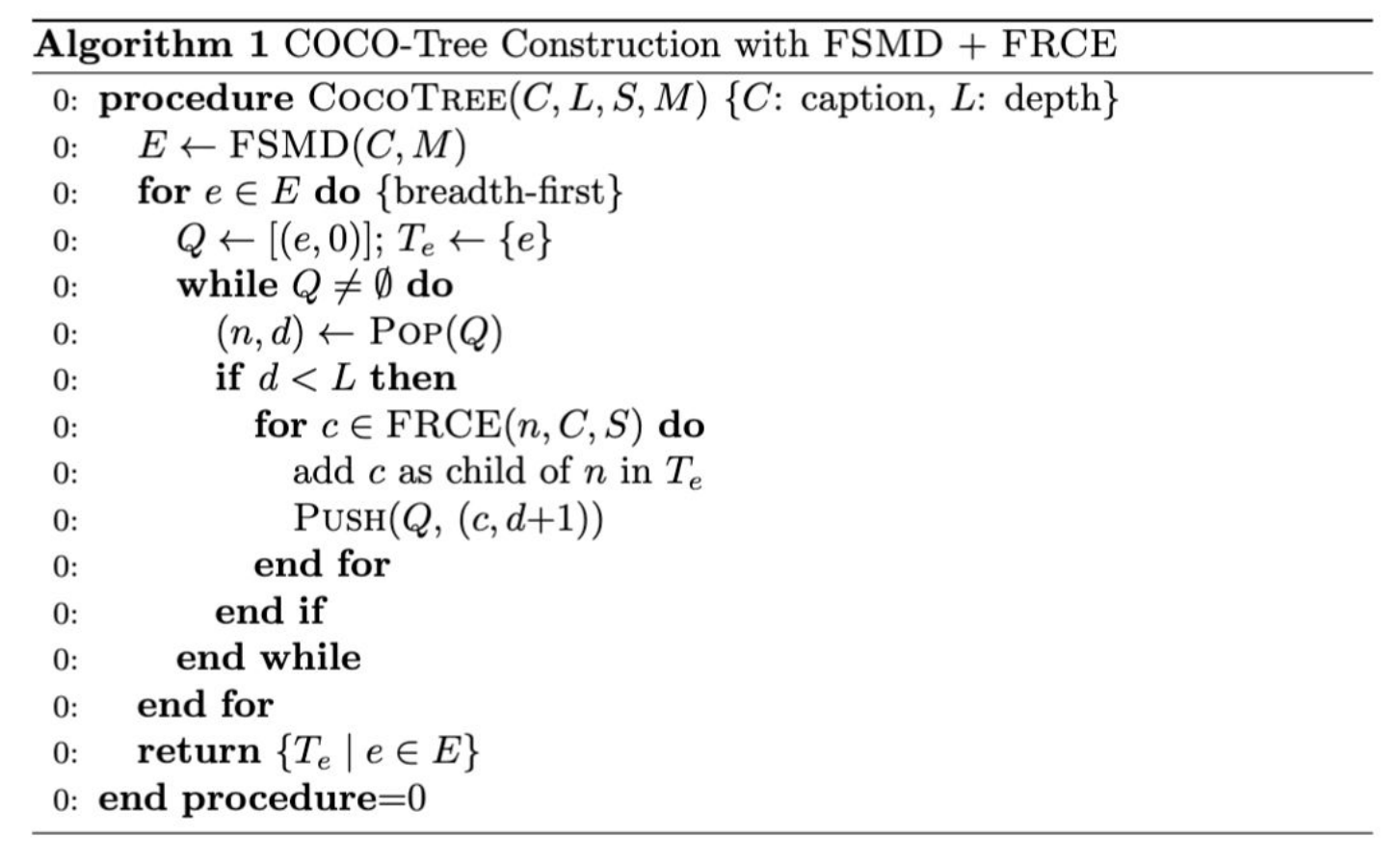}
    \caption{Pseudo-code for concept-tree generation using COCO-Tree.}
    \label{fig:algo}
\end{figure*}

We provide the pseudo-code for generating concept trees in Algorithm~\ref{fig:algo}.

\section{Statistical Significance Testing for COCO-Tree}
We run two statistical significance tests to ascertain the improvements by COCO-Tree over all baselines. We report the Wilcoxon Signed Rank test across all models as compared to baselines in Table~\ref{tab:wilcoxon}.
\begin{table}[h]
\centering
\setlength{\tabcolsep}{8pt}
\renewcommand{\arraystretch}{1.1}
\resizebox{0.5\textwidth}{!}{
\begin{tabular}{lccc}
\toprule
\textbf{Model} & $\bm{\Delta}$\textbf{ Mean (pp)} & \textbf{Wilcoxon $W$} & \textbf{$p$-value} \\
\midrule
LLaVA-1.5-7B         & \phantom{$-$}\textbf{+8.09} & 0 & 0.002$^{\ast\ast}$ \\
LLaVA-1.6-7B         & \phantom{$-$}\textbf{+6.34} & 1 & 0.004$^{\ast\ast}$ \\
Qwen-7B              & \phantom{$-$}+2.76          & 6 & 0.027$^{\ast}$     \\
InternVL-8B          & \phantom{$-$}+1.60          & 0 & 0.012$^{\ast}$     \\
InstructBLIP-XXL     & \phantom{$-$}\textbf{+5.36} & 0 & 0.002$^{\ast\ast}$ \\
LLaVA-1.5            & \phantom{$-$}+4.88          & 0 & 0.002$^{\ast\ast}$ \\
LLaVA-1.6            & \phantom{$-$}\textbf{+6.47} & 0 & 0.002$^{\ast\ast}$ \\
\bottomrule
\end{tabular}}
\caption{Wilcoxon signed-rank test comparing \textbf{COCO-Tree (Beam)} to \textbf{CCoT} for each VLM, $^{\ast\ast}$\,significant at $p<0.01$.}
\label{tab:wilcoxon}
\end{table}

\section{Additional Results on sub-sets}
We report the results of two VLMs - LLaVA-1.5-7b and InstructBLIP-XXL on the labeled sub-sets of the Winoground dataset in Table~\ref{tab:results-subsets}. We observe a win rate of 100\% on Winoground and 86\% on SugarCrepe as shown in Table~\ref{tab:sugar}.

\begin{table*}[t]
\centering
\resizebox{\textwidth}{!}{%
\begin{tabular}{l|c|ccccccccc|cccccccc}
\hline
&  & \multicolumn{3}{c|}{\textbf{Object}} & \multicolumn{3}{c|}{\textbf{Relation}} & \multicolumn{3}{c|}{\textbf{Both}} & \multicolumn{3}{c|}{\textbf{Symbolic}} & \multicolumn{3}{c}{\textbf{Pragmatics}} \\ 
\hline
\textbf{Model} & \textbf{Method} &\textbf{Text} & \textbf{Image} & \textbf{Group} & \textbf{Text} & \textbf{Image} & \textbf{Group} & \textbf{Text} & \textbf{Image} & \textbf{Group} & \textbf{Text} & \textbf{Image} & \textbf{Group} & \textbf{Text} & \textbf{Image} & \textbf{Group} \\ 
\hline
\textbf{LLaVA-1.5-7b}  & VQAScore & 45.39  & 46.10  & 29.08  & 42.06  &  42.49 & 28.76 & 61.54  & 42.31  & 34.62  & 39.29 & 32.14 & 17.86 & 70.59 & 41.18 & 35.29 \\
\textbf{LLaVA-1.5-7b}  &  CCoT  & 46.49  & 46.10  & 31.21  & 42.06  &  42.49 & 28.76 & 61.54  & 42.31  & 34.62  & 39.29 & 32.14 & 17.86 & 70.59 & 41.18 & 35.29 \\
\textbf{LLaVA-1.5-7b}   & COCO-Tree  & \bf 48.23 & \bf 46.10 & \bf 33.33 & \bf 48.07 & \bf 45.06 & \bf 36.05 & \bf 61.54 & \bf 46.15 & \bf 34.62 & \bf 57.14 & \bf 46.43 & \bf 35.71 & \bf 70.59 & \bf 41.18 & \bf 35.29 \\ 
\hline
\textbf{InstructBLIP-XXL} & VQAScore  & 44.68  & 49.64  & 32.62  & 36.05  &  34.76 & 22.75 & 73.08  & 53.85  & 46.15  & 35.71 & 32.14 & 25.00 & 29.41 & 41.18 & 23.53 \\
\textbf{InstructBLIP-XXL} & CCoT & 44.68  & 49.64  & 32.62  & 36.05  &  34.76 & 22.75 & 73.08  & 53.85  & 46.15  & 35.71 & 32.14 & 25.00 & 29.41 & 41.18 & 23.53 \\
\textbf{InstructBLIP-XXL} & COCO-Tree & \bf 46.81 & \bf 48.94 & \bf 39.72 & \bf 48.07 & \bf 46.35 & \bf 38.20 & \bf 73.08 & \bf 53.85 & \bf 46.15 & \bf 57.14 & \bf 50.00 & \bf 46.43 & 29.41 & 41.18 & 23.53  \\ 
\hline
\end{tabular}
}
\caption{Comparison of Compositionality task performance in subsets of the Winoground dataset on LLava-1.5-7b and InstructBLIP-XXL. COCO-tree gets a win rate of 100\%.}
\label{tab:results-subsets}
\end{table*}



\begin{figure*}[t]
\centering
\includegraphics[width=0.75\textwidth]{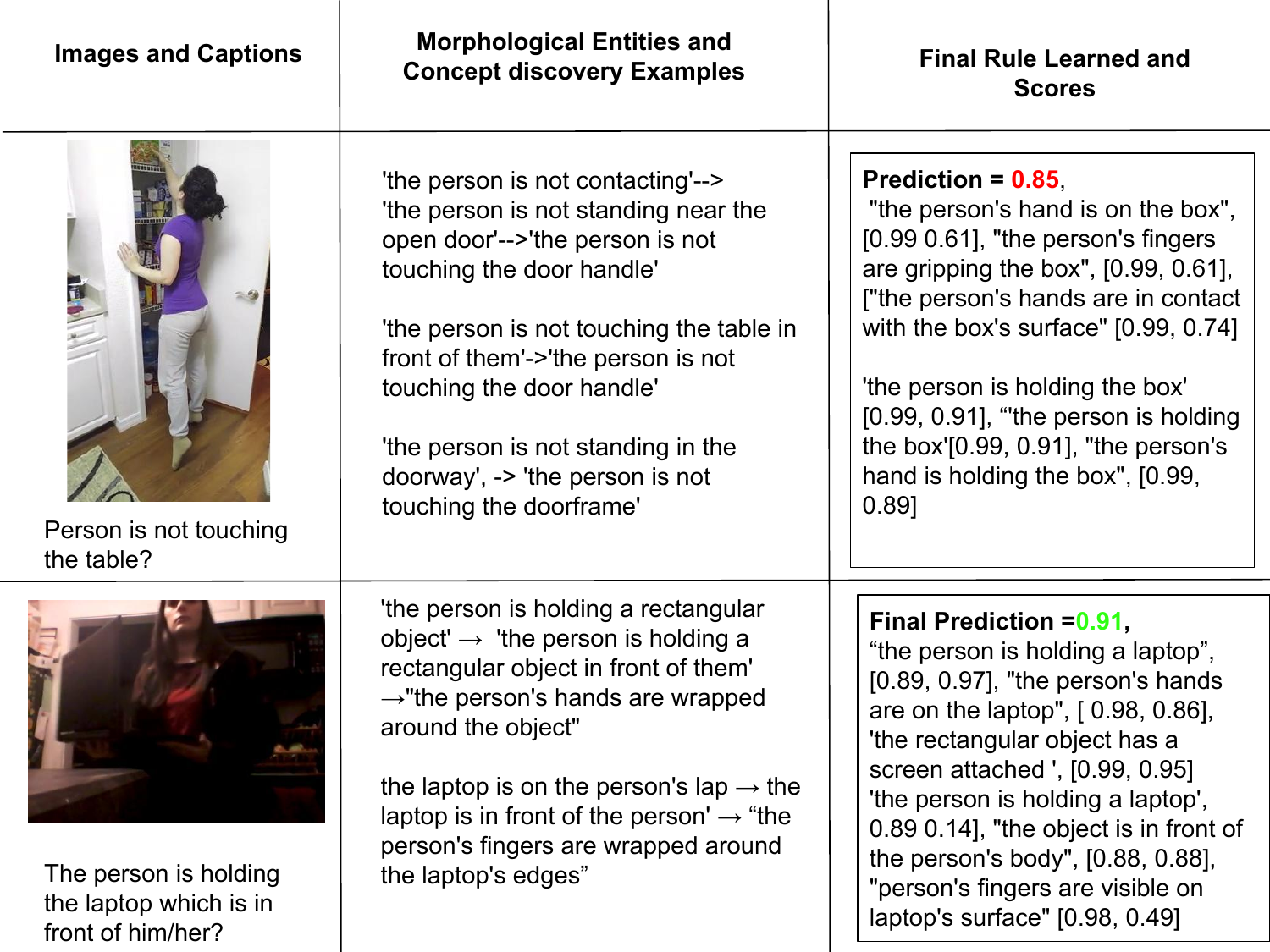}
\caption{Examples demonstrating candidate tree paths and final neurosymbolic rules.}
\label{fig:eg1}
\includegraphics[width=0.75\textwidth]{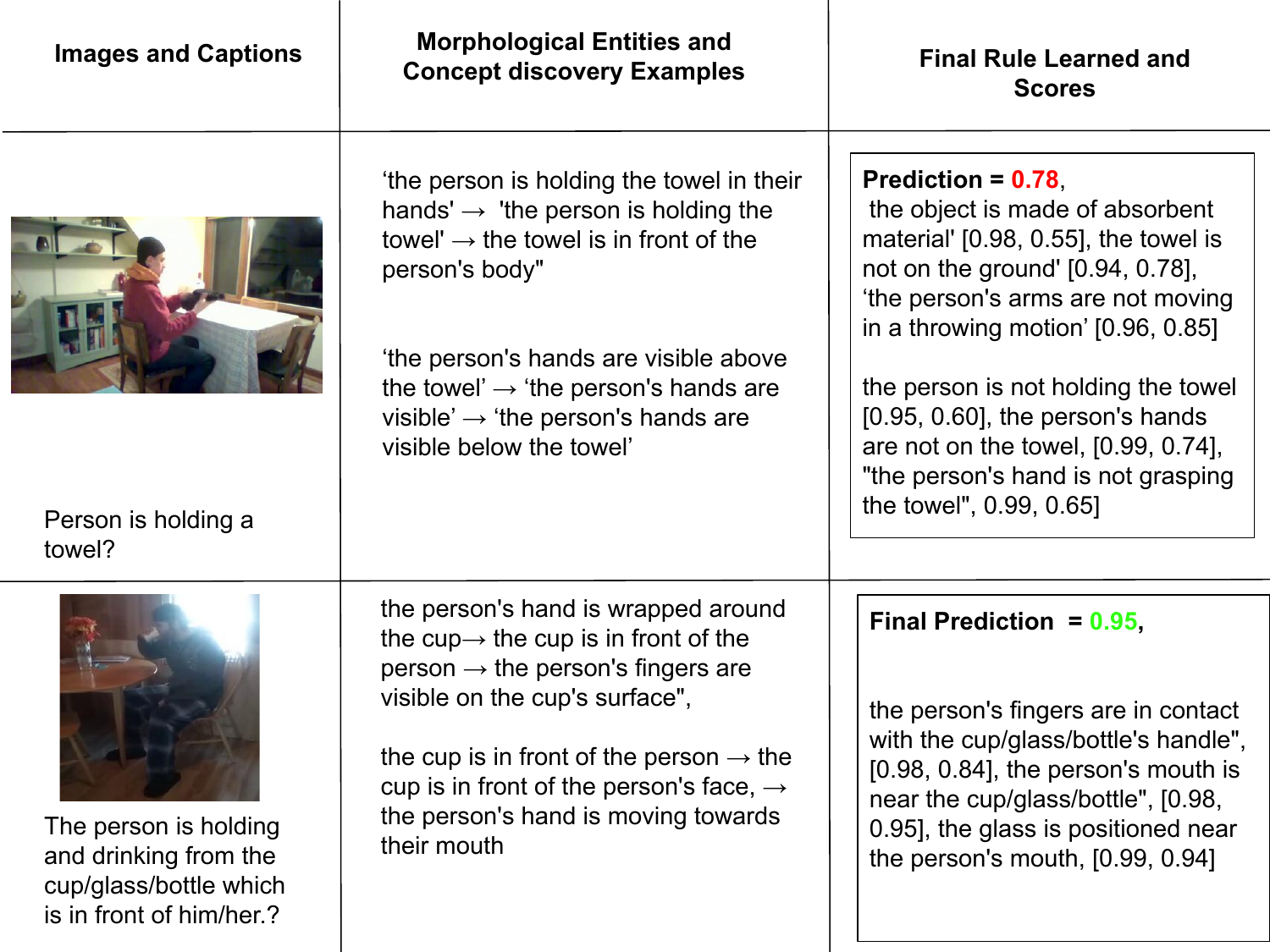}
\caption{Examples demonstrating candidate tree paths and final neurosymbolic rules.}
\label{fig:eg2}
\end{figure*}

\section{Additional Visual Results}
Figures~\ref{fig:eg1} and \ref{fig:eg2} show additional visual examples demonstrating the candidate rules.

\end{document}